\documentclass[letter]{ieice}
\usepackage[pdftex]{graphicx,xcolor}
\usepackage[fleqn]{amsmath}
\usepackage{newtxtext}
\usepackage[varg]{newtxmath}

\usepackage{url}
\usepackage{amsfonts}
\usepackage{booktabs}
\usepackage{multirow}
\usepackage{xspace}
\usepackage[subrefformat=parens]{subcaption}
\usepackage{cite}

\field{D}
\title{Evaluating the Stability of Deep Image Quality Assessment\\ With Respect to Image Scaling}
\authorlist{%
\authorentry[tsubota@hal.t.u-tokyo.ac.jp]{Koki Tsubota}{n}{UTokyo}\MembershipNumber{}
\authorentry[hiroaki.akutsu.cs@hitachi.com]{Hiroaki Akutsu}{m}{Hitachi}\MembershipNumber{1910322}
\authorentry[aizawa@hal.t.u-tokyo.ac.jp]{Kiyoharu Aizawa}{f}{UTokyo}\MembershipNumber{8210642}
}
\affiliate[UTokyo]{The authors are with the Dept. of Information and Communication Engineering, The University of Tokyo, Tokyo, Japan.}
\affiliate[Hitachi]{The author is with Center for Technology Innovation - Digital Platform, Research \& Development Group, Hitachi, Ltd., Kanagawa, Japan.}

\received{2022}{3}{13}
\revised{2022}{3}{13}

\makeatletter
\DeclareRobustCommand\onedot{\futurelet\@let@token\@onedot}
\def\@onedot{\ifx\@let@token.\else.\null\fi\xspace}

\def\etal{\emph{et al}\onedot}
\makeatother

\begin{document}
\maketitle
\begin{summary}
    Image quality assessment (IQA) is a fundamental metric for image processing tasks (e.g., compression).
    With full-reference IQAs, traditional IQAs, such as PSNR and SSIM, have been used.
    Recently, IQAs based on deep neural networks (deep IQAs), such as LPIPS and DISTS, have also been used.
    It is known that image scaling is inconsistent among deep IQAs, as some perform down-scaling as pre-processing, whereas others instead use the original image size.
    In this paper, we show that the image scale is an influential factor that affects deep IQA performance.
    We comprehensively evaluate four deep IQAs on the same five datasets, and the experimental results show that image scale significantly influences IQA performance.
    We found that the most appropriate image scale is often neither the default nor the original size, and the choice differs depending on the methods and datasets used.
    We visualized the stability and found that PieAPP is the most stable among the four deep IQAs.
\end{summary}

\section{Introduction}
\vspace{-5pt}
This paper discusses full-reference image quality assessment (IQA), which aims to predict how humans perceive the visual quality of distorted images by comparing them to non-distorted references.
Full-reference IQA provides evaluation criteria for image processing tasks, such as image compression and image restoration.
Recently, the full-reference IQA has also been used as a loss function for such tasks~\cite{MentzerHific,DingComparisonIJCV21}.

Peak signal-to-noise ratio (PSNR) and structural similarity (SSIM) index metrics~\cite{SSIM} are widely used full-reference IQAs.
Despite their popularity, they do not accurately model human perception.
To better model our complex visual system, IQAs using deep neural networks (deep IQAs) have recently been investigated~\cite{LPIPS,PieAPP,DISTS}, achieving a better correlation to human perception.
Specifically, learned perceptual image patch similarity (LPIPS)~\cite{LPIPS}, a deep IQA, is widely used as the evaluation metric in computer vision tasks~\cite{MentzerHific}.

This study examines the image scale used when predicting image quality.
Many traditional IQAs~\cite{SSIM,FSIM,GMSD} down-scale images to a specific size for pre-processing.
However, the process is inconsistent among deep IQAs, as some ~\cite{DISTS,Deepsim} perform down-scaling and others~\cite{journals/tip/BosseMMWS18,LPIPS,PieAPP} use the original size.

In this paper, we consider the image scale as the independent variable and empirically evaluate its effect on IQA performance.
Notably, a few other researchers have investigated this effect.
Joy \etal~\cite{JoyARPN}, for example, studied the effect of image scale on SSIM~\cite{SSIM} and feature similarity index (FSIM) metrics~\cite{FSIM}.
Gao \etal~\cite{Deepsim} used the distance between intermediate representations extracted by a VGG network~\cite{VGG} pre-trained on ImageNet~\cite{ImageNet} for IQA.
They investigated two image sizes: original (e.g., $384\times512$ pixels in the TID2013~\cite{TID2013} dataset) and resized ($224\times224$ pixels), empirically demonstrating that using the resized version improved performance.
However, their investigation was limited to these two sizes.

In our experiments, we evaluate four deep IQAs: Alex-lin and VGG-lin of LPIPS~\cite{LPIPS}, the perceptual image-error assessment through pairwise preference (PieAPP)~\cite{PieAPP}, and deep image structure and texture similarity (DISTS)~\cite{DISTS}.
We apply each of these IQAs to the same five datasets.
Through comprehensive experimentation, we show the effect of image scale on IQA performance.

Our contributions are summarized as follows:
\begin{itemize}
  \item Using the image scale as the independent variable, we evaluate deep IQAs with different image scales and comprehensively measure their effect on performance. To the best of our knowledge, this is the first work to do so.
  \item We empirically show that the image scale greatly affects deep IQA performance and that the most desirable image scale is often neither the default nor the original size; it differs depending on the deep IQAs and the datasets used.
  \item We visualize the stability of the deep IQAs with respect to their scales and performance scores and found that PieAPP is the most stable.
\end{itemize}

\section{Related Works}
\vspace{-5pt}
IQA is a fundamental image processing task that has been studied for several decades.
The simplest IQA is the mean squared error (MSE).
It averages the squared errors of pixel values to obtain pixel-level errors.
PSNR is a different representation of MSE.
SSIM~\cite{SSIM} is also widely used.
It captures the structural similarity of images instead of their pixel-level errors.
FSIM~\cite{FSIM} uses phase congruency and gradient magnitude as features to calculate similarity.
Multi-scale SSIM (MS-SSIM)~\cite{MSSSIM} calculates SSIM at multiple scales by down-scaling the image.

With the development of deep IQAs, several related studies have been published.
Zhang \etal~\cite{LPIPS} constructed a large-scale patch dataset for examining perceptual loss~\cite{conf/eccv/JohnsonAF16}, which is the computed distance between intermediate representations of models pre-trained on ImageNet~\cite{ImageNet}.
They also proposed LPIPS, which performs a weighted summation of the distance across channels.
Prashnani \etal~\cite{PieAPP} presented a pairwise-learning framework using their PieAPP dataset.
Ding \etal~\cite{DISTS} focused on image textures and trained an invariant model under texture shifts.
Their model computes similarity not only in the feature space, but also in the image space to obtain an injective function.
Notably, the image scaling used for these IQAs was inconsistent.

\section{Method}
\vspace{-5pt}
\begin{figure}[t]
   \centering
   \includegraphics[width=\hsize]{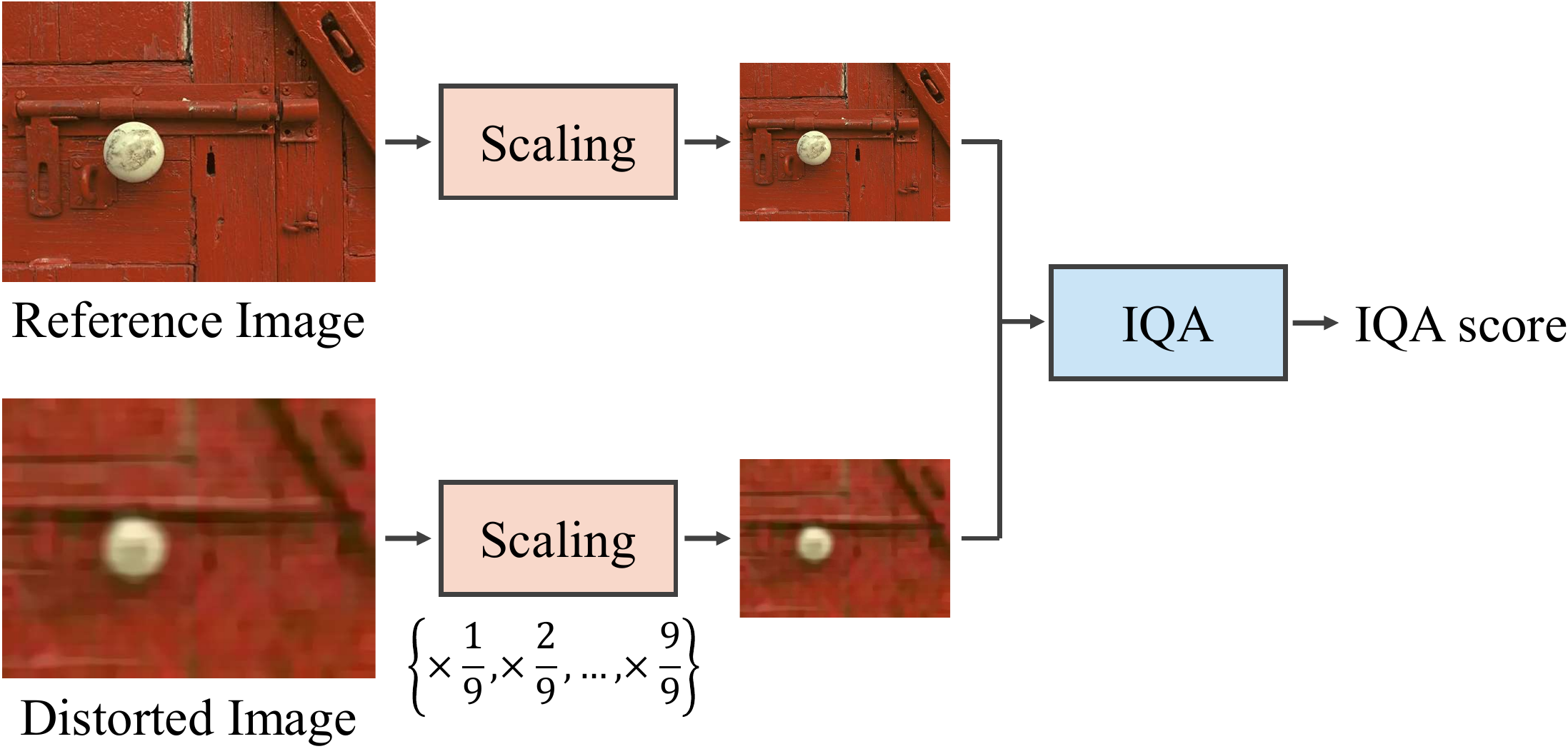}
   \caption{Outline of our image quality assessment (IQA) evaluation procedure.}
   \label{fig:overview}
   \vspace{-5pt}
\end{figure}
We illustrate the flow of our evaluation procedure in Fig.~\ref{fig:overview}.
Given a reference and distorted image pair, we perform down-scaling on both using various scales for pre-processing.
We then apply IQAs to the pair to obtain the IQA score.

We use four deep IQAs (i.e., Alex-lin and VGG-lin of LPIPS~\cite{LPIPS}, PieAPP~\cite{PieAPP}, and DISTS~\cite{DISTS}) for evaluation.
We list the training data and initial weights of these IQAs in Table~\ref{tbl:method}.
PieAPP is initialized with random weights, whereas the others are initialized with ImageNet~\cite{ImageNet} pre-trained weights that are fixed during training.

LPIPS~\cite{LPIPS} extracts intermediate representations from an image pair using a backbone network and computes their weighted distance per channel.
Alex-lin and VGG-lin respectively use AlexNet~\cite{AlexNet} and VGG~\cite{VGG} as the backbone network.
LPIPS does not perform down-scaling during training and inference.

PieAPP~\cite{PieAPP} leverages a VGG-based network architecture on the PieAPP dataset~\cite{PieAPP} using multiple random $64 \times 64$ cropped patches.
During inference, PieAPP is applied to the image patches cropped using sliding window.
PieAPP does not perform down-scaling during training and inference.

DISTS~\cite{DISTS} uses a VGG network as the backbone network.
It calculates the global similarity between intermediate representations of the backbone network extracted from an image pair.
It is applied to scaled image pairs whose shorter side is set to $256$ pixels during training and inference.

\begin{table}[t]
  \centering
  \caption{Deep image quality assessments used in our experiments. IN denotes the ImageNet pre-trained weights and RND denotes random weights.}
  \label{tbl:method}
  \begin{tabular}{lcc}
    \toprule
    Method & Training Data & Initial Weights \\
    \midrule
    LPIPS (Alex-lin)~\cite{LPIPS} & BAPPS~\cite{LPIPS} & IN\\
    LPIPS (VGG-lin)~\cite{LPIPS} & BAPPS~\cite{LPIPS} & IN\\
    PieAPP~\cite{PieAPP} & PieAPP~\cite{PieAPP} & RND\\
    DISTS~\cite{DISTS} & KADID-10k~\cite{KADID10K}, DTD~\cite{DTD} & IN \\
    \bottomrule
  \end{tabular}
  \vspace{-5pt}
\end{table}

\section{Experiments}
\vspace{-5pt}
\subsection{Experimental Settings}
\vspace{-5pt}
\begin{table}[t]
  \centering
  \caption{Image quality assessment datasets used in our experiments. ``Ref.'' denotes reference images, and ``Dis.'' denotes distorted images.}
  \label{tbl:dataset}
  \begin{tabular}{lcccc}
    \toprule
    Dataset & Image size & \# Ref. & \# Dis. & Metric \\
    \midrule
    TID2013~\cite{TID2013} & $384\times512$ & 25 & 3,000 & SRCC \\
    PieAPP~\cite{PieAPP} & $256\times256$ & 40 & 600 & SRCC \\
    KADID-10k~\cite{KADID10K} & $384\times512$ & 81 & 10,125 & SRCC \\
    PIPAL~\cite{PIPAL} & $288\times288$ & 200 & 23,200 & SRCC \\
    CLIC2021~\cite{CLIC21} & $768 \times 768$ & 310 & 2,405 & ACC \\
    \bottomrule
  \end{tabular}
  \vspace{-5pt}
\end{table}

We evaluated the four deep IQAs using the same five datasets: TID2013~\cite{TID2013}, PieAPP~\cite{PieAPP}, KADID-10k~\cite{KADID10K}, PIPAL~\cite{PIPAL}, and CLIC2021 validation dataset for perceptual metrics (CLIC2021)~\cite{CLIC21}.
For reference, we evaluated three traditional methods: PSNR, SSIM~\cite{SSIM}, and MS-SSIM~\cite{MSSSIM}.
In Table~\ref{tbl:dataset}, we list the properties of these datasets (i.e., image size, number of images, and metric used).
PieAPP and PIPAL datasets comprise patch images, whereas TID2013, KADID-10k, and CLIC2021 datasets comprise whole images.
The CLIC2021 dataset contains higher-resolution images than the other datasets.
We used images of uniform size from the CLIC2021 dataset by removing five reference images and their corresponding distorted images, whose sizes were not $768 \times 768$ pixels.

The TID2013 and KADID-10k datasets contain distortions applied by traditional operations, such as color change and JPEG compression.
PieAPP and PIPAL datasets contain distortions applied by traditional operations as well as obtained by algorithm outputs such as super-resolution.
The PIPAL dataset includes the output of a trained model with adversarial loss.

\begin{table*}[t]
  \centering
  \caption{Performance scores of deep image quality assessment (IQA) metrics when changing the image scale for each IQA dataset. The underline denotes the performance of the default scale, and the bold font denotes the best performance.}
  \label{tbl:result_iqa}
  \begin{tabular}{llcccccccccc}
    \toprule
    \multirow{2}{*}{Method} & \multirow{2}{*}{Dataset}
    & \multicolumn{10}{c}{Image Scale}\\
    & & $\times1/9$ & $\times2/9$ & $\times3/9$ & $\times4/9$ & $\times1/2$ & $\times5/9$ & $\times6/9$ & $\times7/9$ & $\times8/9$ & original size\\
    \midrule
    \multirow{5}{*}{PSNR}
    & TID2013 &  0.553 &  0.606 &  0.643 &  0.665 &  0.669 &  0.685 &  0.700 &  0.710 &  \textbf{0.715} &  \underline{0.687} \\
    & PieAPP &  0.221 &  0.234 &  0.253 &  0.266 &  0.268 &  0.272 &  0.275 &  \textbf{0.279} &  \textbf{0.279} &  \underline{0.268} \\
    & KADID-10k &  0.388 &  0.457 &  0.512 &  0.548 &  0.555 &  0.579 &  0.604 &  0.624 &  0.637 &  \underline{\textbf{0.676}} \\
    & PIPAL &  0.279 &  0.415 &  0.494 &  0.521 &  \textbf{0.525} &  0.513 &  0.501 &  0.484 &  0.471 &  \underline{0.407} \\
    & CLIC2021 &  0.384 &  0.387 &  0.394 &  0.398 &  0.399 &  0.402 &  0.403 &  0.407 &  0.409 &  \underline{\textbf{0.427}} \\
    \midrule
    \multirow{5}{*}{SSIM~\cite{SSIM}}
    & TID2013 &  0.624 &  0.732 &  \textbf{0.762} &  0.756 &  \underline{0.752} &  0.738 &  0.716 &  0.686 &  0.657 &  0.554 \\
    & PieAPP &  0.187 &  0.230 &  0.277 &  0.309 &  0.316 &  0.326 &  0.340 &  0.345 & \textbf{0.348} &  \underline{0.321} \\
    & KADID-10k &  0.508 &  0.634 &  0.698 &  0.719 &  \underline{0.722} &  \textbf{0.723} &  0.717 &  0.704 &  0.688 &  0.633 \\
    & PIPAL &  0.286 &  0.417 &  0.498 &  0.545 &  0.556 &  0.556 &  \textbf{0.557} &  0.550 &  0.541 &  \underline{0.498} \\
    & CLIC2021 &  0.357 &  0.358 &  \underline{0.368} &  0.373 &  0.377 &  0.384 &  0.391 &  0.401 &  0.407 &  \textbf{0.429} \\
    \midrule
    \multirow{5}{*}{MS-SSIM~\cite{MSSSIM}}
    & TID2013 & - &    - &    - &  0.793 &  0.803 &  0.813 &  0.821 &  \textbf{0.823} &  0.820 &  \underline{0.798} \\
    & PieAPP & - &    - &    - &    - &    - &    - &  0.296 &  0.305 &  0.311 &  \underline{\textbf{0.315}} \\
    & KADID-10k & - &    - &    - &  0.725 &  0.741 &  0.760 &  0.783 &  0.796 &  \textbf{0.803} &  \underline{0.802} \\
    & PIPAL & - &    - &    - &    - &    - &    - &  0.522 &  0.541 &  0.547 &  \underline{\textbf{0.552}} \\
    & CLIC2021 & - &  0.361 &  0.362 &  0.361 &  0.363 &  0.366 &  0.370 &  0.373 &  0.376 &  \underline{\textbf{0.388}} \\
    \midrule
    \multirow{5}{*}{LPIPS (Alex-lin)~\cite{LPIPS}}
    & TID2013 &  0.590 &  0.724 &  0.807 &  0.834 &  \textbf{0.841} & \textbf{0.841} &  0.833 &  0.817 &  0.796 &  \underline{0.744}\\
    & PieAPP &    - &  0.331 &  0.418 &  0.496 &  0.516 & 0.544 &  0.573 &  \textbf{0.578} &  0.571 &  \underline{0.550}\\
    & KADID-10k & 0.528 &  0.696 &  0.802 &  0.856 & 0.869 & 0.883 &  \textbf{0.891} &  0.886 &  0.870 &  \underline{0.822}\\
    & PIPAL & 0.305 &  0.471 &  0.594 &  0.675 & \textbf{0.684} & \textbf{0.684} &  0.670 &  0.650 &  0.627 &  \underline{0.585}\\
    & CLIC2021 &  0.645 &  0.692 &  0.732 &  0.746 & 0.753 & \textbf{0.760} &  \textbf{0.760} &  0.755 &  0.750 &  \underline{0.737}\\
    \midrule
    \multirow{5}{*}{LPIPS (VGG-lin)~\cite{LPIPS}}
    & TID2013 & 0.699 &  0.819 &  \textbf{0.842} &  0.826 & 0.816 & 0.801 &  0.774 &  0.747 &  0.722 &  \underline{0.670}\\
    & PieAPP & 0.240 &  0.326 &  0.398 &  0.448 & 0.461 & 0.477 &  0.495 &  0.500 &  \textbf{0.503} &  \underline{0.492}\\
    & KADID-10k & 0.610 &  0.778 &  0.837 &  \textbf{0.848} & 0.844 & 0.840 &  0.820 &  0.797 &  0.774 &  \underline{0.720}\\
    & PIPAL & 0.286 &  0.433 &  0.533 &  0.595 & 0.603 & \textbf{0.611} &  0.609 &  0.601 &  0.592 &  \underline{0.573}\\
    & CLIC2021 & 0.669 &  0.718 &  0.737 &  0.742 & 0.745 & \textbf{0.746} &  0.745 &  0.741 &  0.740 &  \underline{0.745}\\
    \midrule
    \multirow{5}{*}{PieAPP~\cite{PieAPP}}
    & TID2013 & - &  0.679 &  0.783 &  0.823 & 0.837 & 0.848 &  0.863 &  0.871 &  \textbf{0.872} &  \underline{0.836}\\
    & PieAPP & - & - &  0.463 &  0.601 & 0.664 & 0.672 &  0.732 &  0.773 &  \textbf{0.796} &  \underline{\textbf{0.796}}\\
    & KADID-10k & - &  0.578 &  0.701 &  0.768 & 0.794 & 0.815 &  0.842 &  0.855 &  0.861 &  \underline{\textbf{0.865}}\\
    & PIPAL & - &  0.295 &  0.422 &  0.548 & 0.591 & 0.638 &  0.690 &  0.705 &  \textbf{0.712} &  \underline{0.698}\\
    & CLIC2021 & 0.537 &  0.612 &  0.671 &  0.699 & 0.701 & 0.705 &  0.724 &  0.733 &  0.747 &  \underline{\textbf{0.760}}\\
    \midrule
    \multirow{5}{*}{DISTS~\cite{DISTS}}
    & TID2013 & 0.673 &  0.803 &  0.850 & \textbf{0.857} & 0.855 & 0.844 &  \underline{0.830} &  0.807 &  0.784 &  0.708\\
    & PieAPP & 0.266 &  0.385 &  0.535 &  0.644 & 0.699 & 0.731 &  0.762 &  \textbf{0.770} &  0.756 &  \underline{0.693}\\
    & KADID-10k & 0.586 &  0.764 &  0.847 &  0.878 & 0.885 & \textbf{0.890} &  \underline{0.887} &  0.876 &  0.862 &  0.814\\
    & PIPAL & 0.297 &  0.455 &  0.580 &  0.646 & 0.651 & \textbf{0.660} &  0.652 &  0.637 &  \underline{0.624} &  0.579\\
    & CLIC2021 & 0.664 &  0.716 &  \underline{0.742} &  0.754 & 0.758 & \textbf{0.759} &  0.756 &  0.757 &  0.753 &  0.748\\
    \bottomrule
  \end{tabular}
  \vspace{-5pt}
\end{table*}

\begin{figure*}[t]
  \begin{minipage}[t]{0.24\hsize}
    \centering
    \includegraphics[width=\hsize]{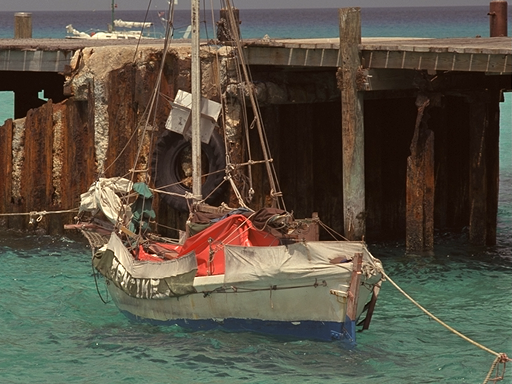}
    \captionsetup{justification=centering}
    \subcaption*{distortion type\\
    annotated IQA score$\uparrow$\\
    LPIPS (Alex-lin) default / best$\downarrow$\\
    LPIPS (VGG-lin) default / best$\downarrow$\\
    PieAPP default / best$\downarrow$\\
    DISTS default / best$\downarrow$}
  \end{minipage}
  \begin{minipage}[t]{0.24\hsize}
    \centering
    \includegraphics[width=\hsize]{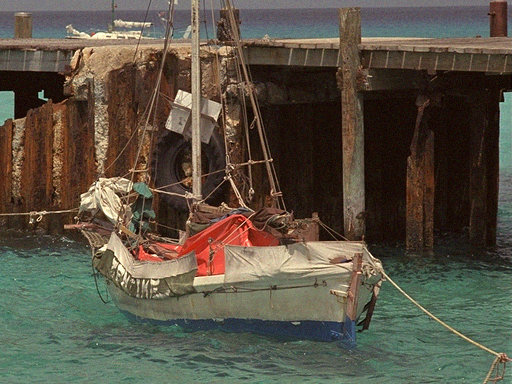}
    \captionsetup{justification=centering}
    \subcaption*{additive Gaussian noise\\
    5.541 (1)\\
    0.034 (1) / \underline{0.004 (1)}\\
    0.106 (2) / \underline{0.008 (1)}\\
    0.451 (1) / \underline{0.304 (1)}\\
    0.024 (1) / \underline{0.014 (1)}}
  \end{minipage}
  \begin{minipage}[t]{0.24\hsize}
    \centering
    \includegraphics[width=\hsize]{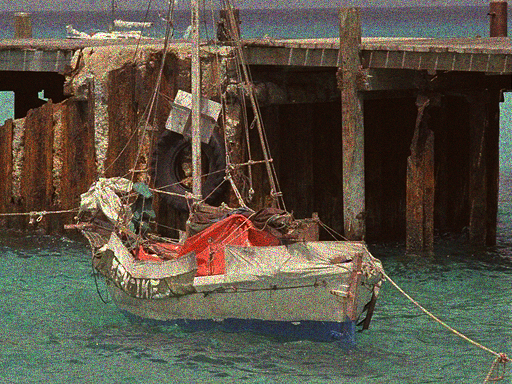}
    \captionsetup{justification=centering}
    \subcaption*{multiplicative Gaussian noise\\
    3.971 (2)\\
    0.189 (3) / \underline{0.030 (2)}\\
    0.285 (3) / \underline{0.037 (2)}\\
    1.436 (3) / \underline{0.891 (2)}\\
    0.108 (3) / \underline{0.067 (2)}}
  \end{minipage}
  \begin{minipage}[t]{0.24\hsize}
    \centering
    \includegraphics[width=\hsize]{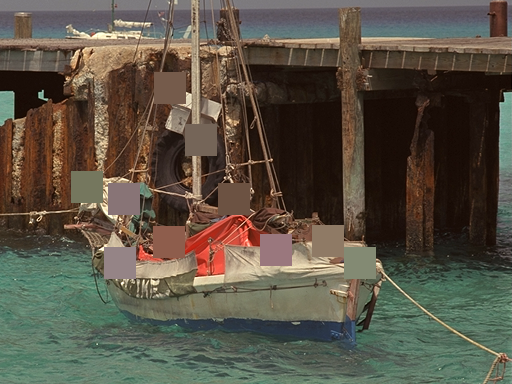}
    \captionsetup{justification=centering}
    \subcaption*{local block-wise distortions\\
    3.189 (3)\\
    0.067 (2) / \underline{0.054 (3)}\\
    0.065 (1) / \underline{0.071 (3)}\\
    1.012 (2) / \underline{1.250 (3)}\\
    0.072 (2) / \underline{0.084 (3)}}
  \end{minipage}
  \caption{Qualitative results. The images in the far-left column are reference images. The remaining columns contain samples of distorted images sorted by annotated IQA scores. ``Default'' denotes the default scale, and ``best'' denotes the best scale for each IQA. The value in parentheses denotes the ranking of the annotated or predicted image quality. The underline denotes that the predicted IQA scores are in the correct order.}
  \label{fig:qualitative}
  \vspace{-10pt}
\end{figure*}

The TID2013, PieAPP, KADID-10k, and PIPAL datasets comprise pairs of reference and distorted images.
Because each pair is annotated with its relative image quality in the dataset, we used Spearman's rank correlation coefficients (SRCC) as the evaluation metric for these datasets.
The CLIC2021 dataset comprises sets containing a reference image and two distorted images.
Because each set was annotated to identify the distorted image with better quality, we used accuracy (ACC) as the evaluation metric for CLIC2021 dataset.

We down-scaled the images using 10 different ratios: $\{\times 1/9, \times 2/9, \times 3/9, \times 4/9, \times 1/2, \times 5/9, \times 7/9, \times 8/9, $ and $\times 9/9\}$.
$\times 9/9$ is the original size.
These scales included the default scale for each method in each dataset.
We used bilinear down-scaling and evaluated the IQA performance when the image size was larger than the minimum size for each IQA model.

\subsection{Experimental Results}
\vspace{-5pt}
We present the experimental results in Table~\ref{tbl:result_iqa}.
The results show that the appropriate image scale differs depending on the IQA and evaluation dataset.

Interestingly, the best image scale, where the SRCC or ACC are the best, differs from the default scale and the original size in most cases.
For example, the best image scale for LPIPS is $\times 1/2, \times 7/9, \times 6/9, \times 1/2,$ and $\times 5/9$ for TID2013, PieAPP, KADID-10k, PIPAL, and CLIC2021, respectively.
These scales differ from the original image size and the default scale.

Down-scaling improves the score in most cases, but discrepancies tends to depend on the method and dataset.
Specifically, the difference in LPIPS is large.
The SRCC of LPIPS (VGG-lin) when using the TID2013 dataset differs from the default scale by 0.172.

\begin{figure}[t]
  \begin{minipage}[t]{0.48\hsize}
    \centering
    \includegraphics[width=\hsize]{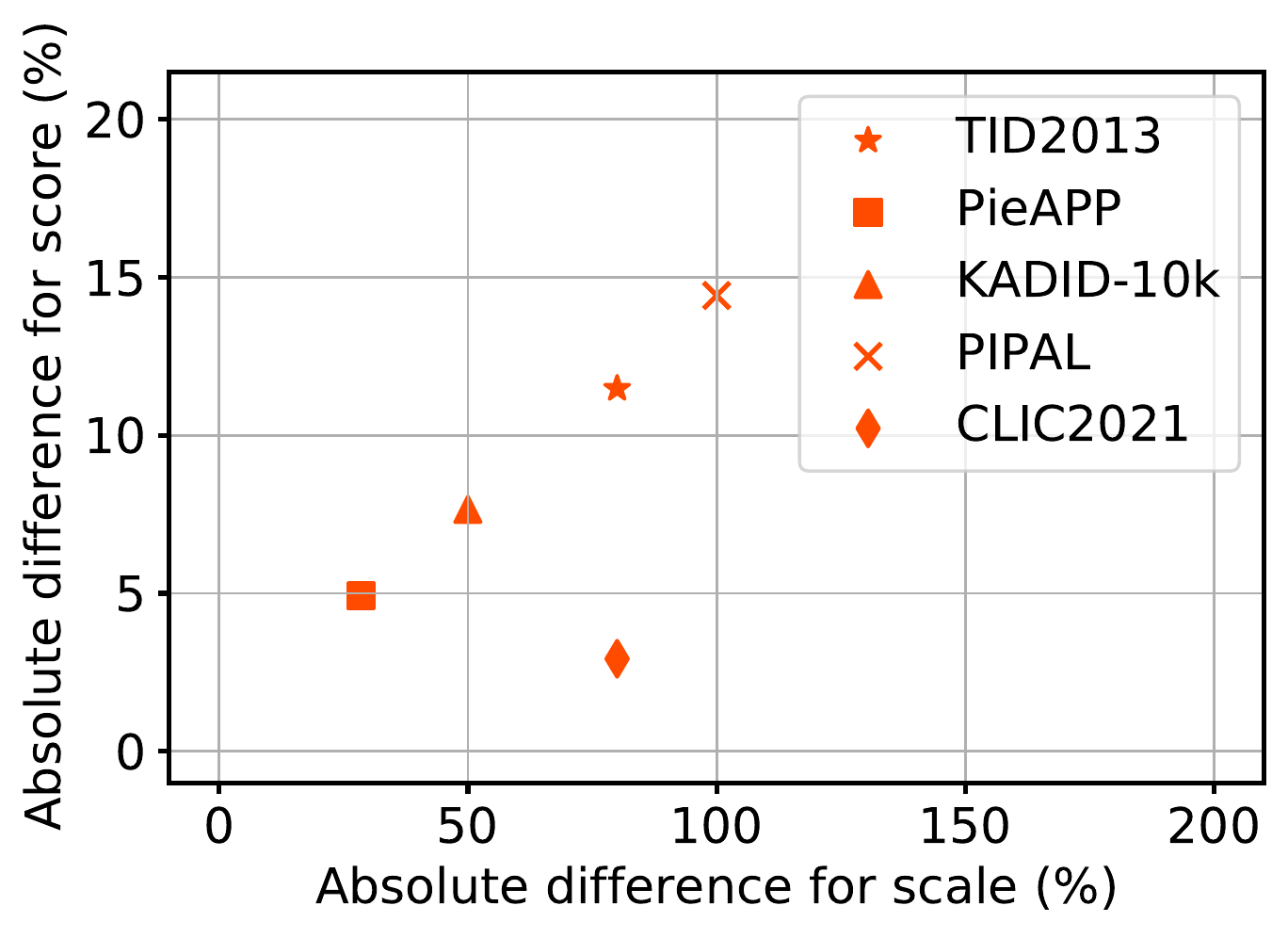}
    \subcaption*{LPIPS (Alex-lin)~\cite{LPIPS}}
  \end{minipage}
  \begin{minipage}[t]{0.48\hsize}
    \centering
    \includegraphics[width=\hsize]{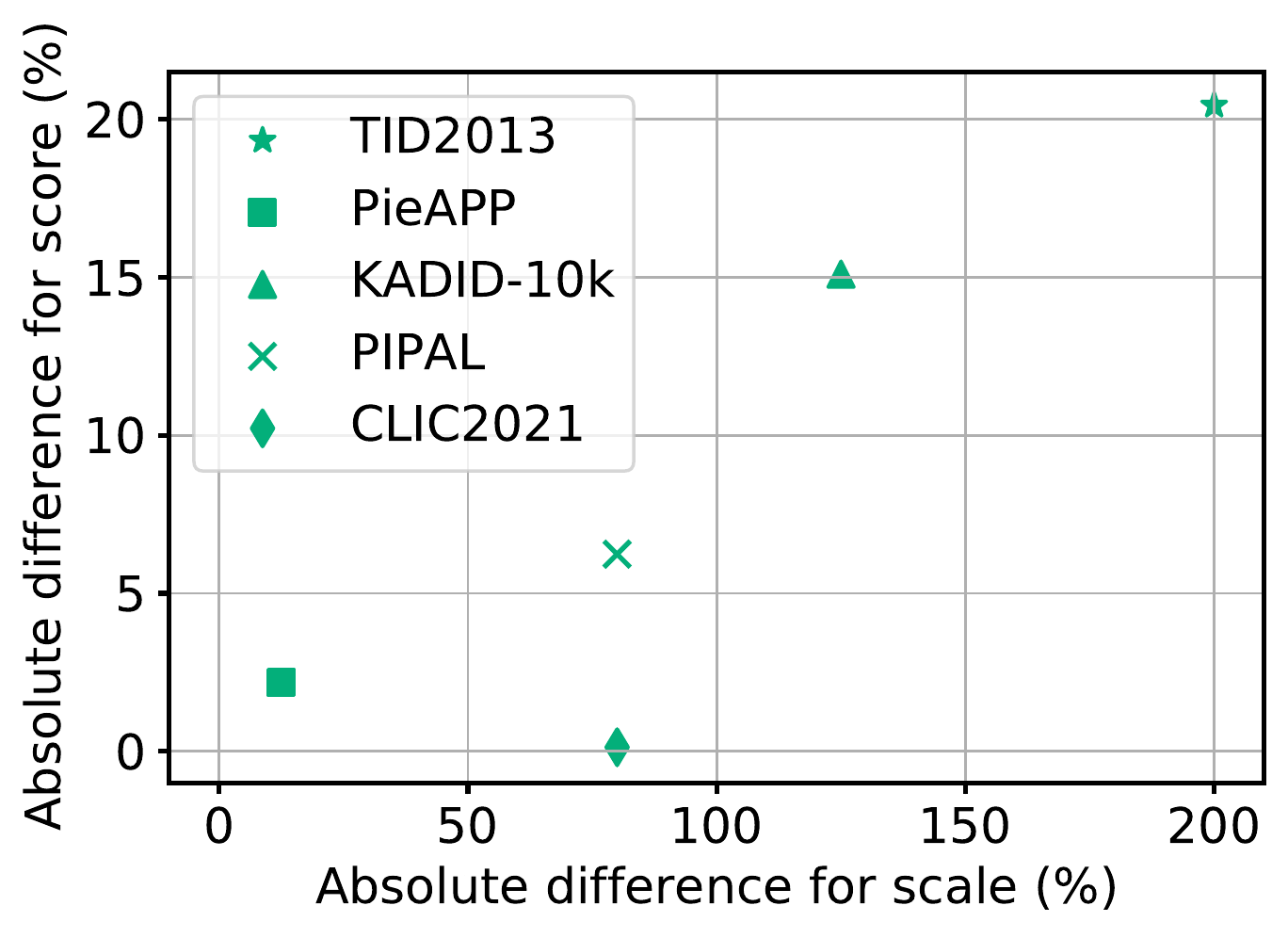}
    \subcaption*{LPIPS (VGG-lin)~\cite{LPIPS}}
  \end{minipage}
  
  \begin{minipage}[t]{0.48\hsize}
    \centering
    \includegraphics[width=\hsize]{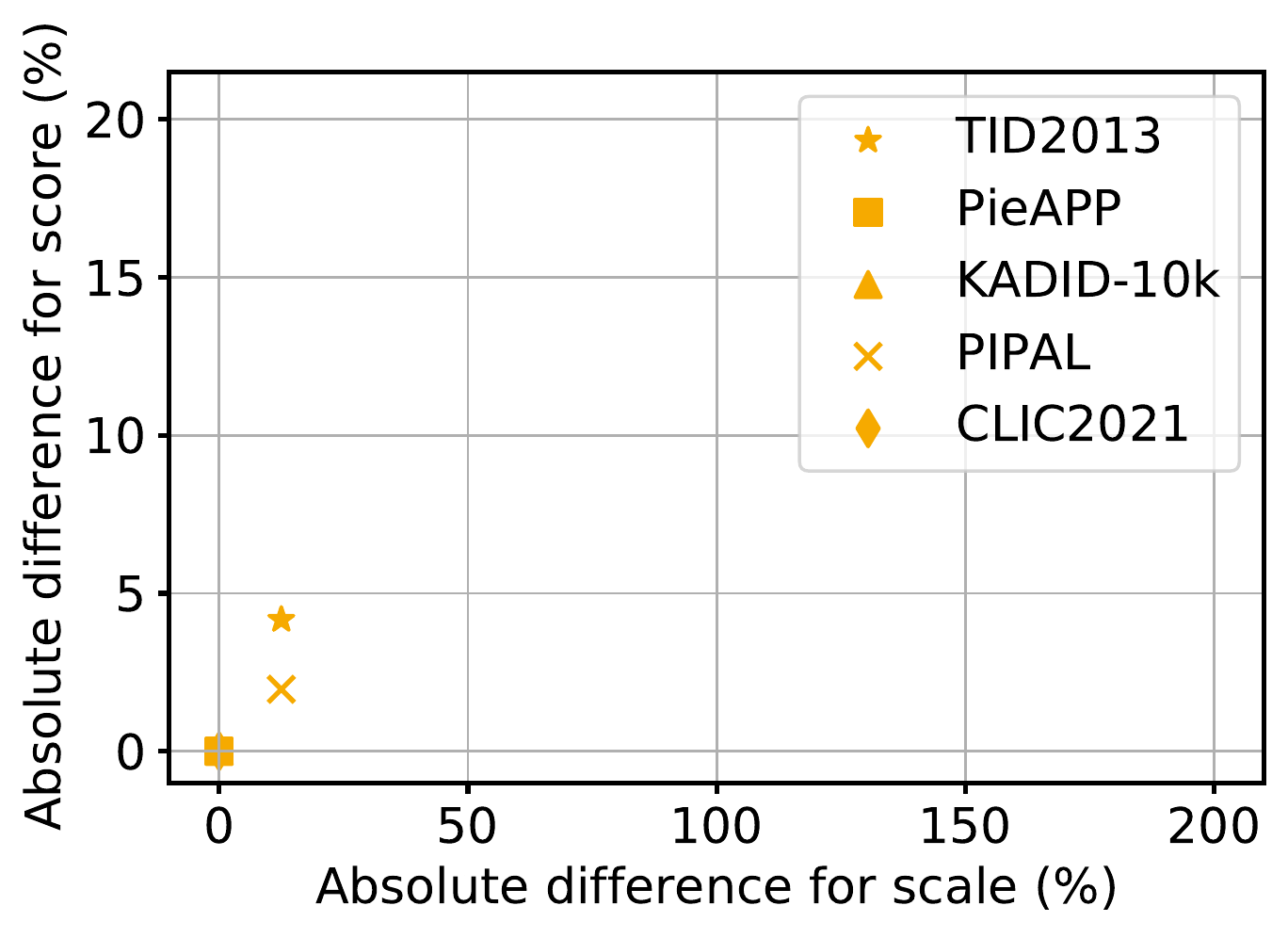}
    \subcaption*{PieAPP~\cite{PieAPP}}
  \end{minipage}
  \begin{minipage}[t]{0.48\hsize}
    \centering
    \includegraphics[width=\hsize]{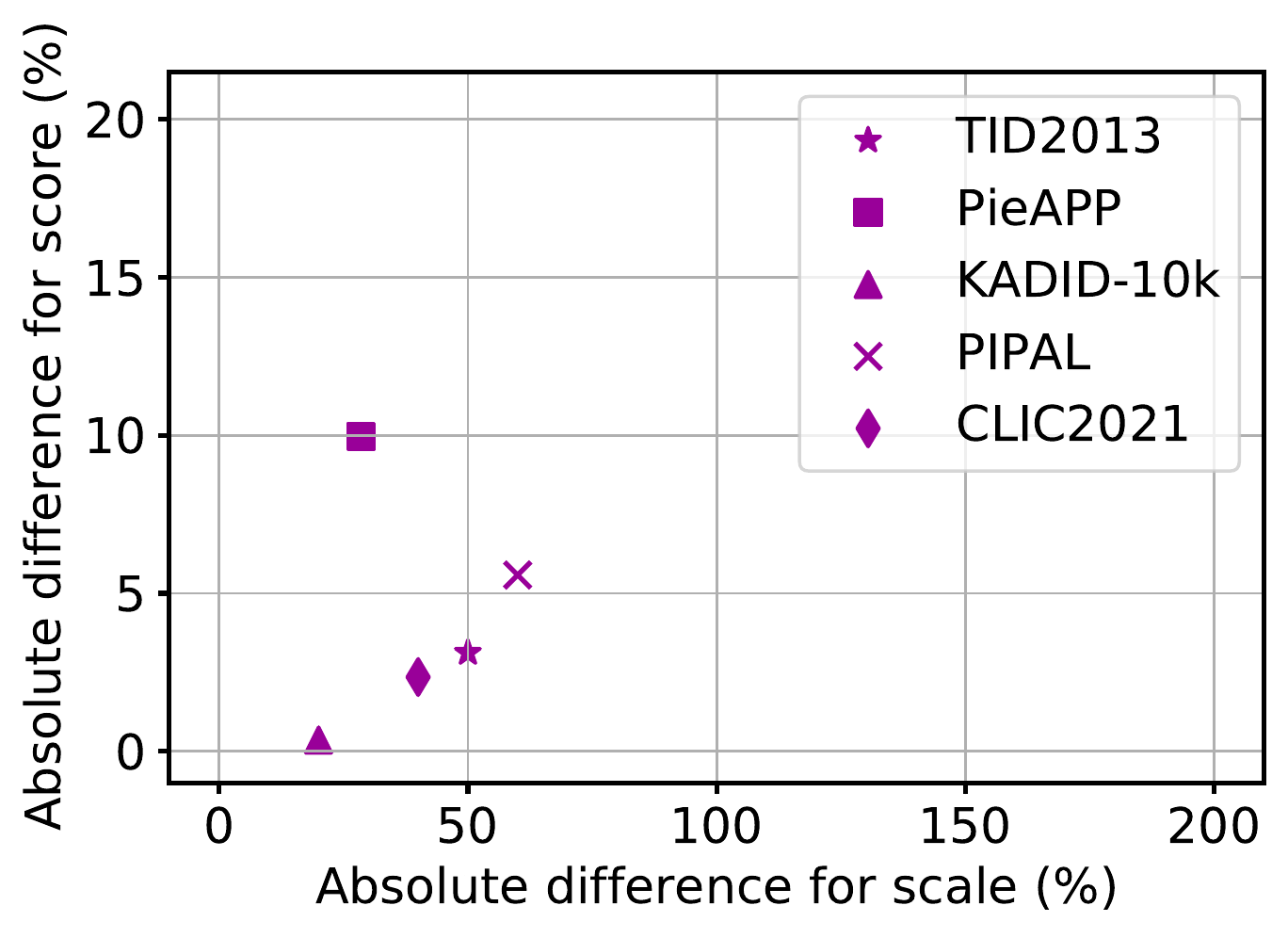}
    \subcaption*{DISTS~\cite{DISTS}}
  \end{minipage}
  \caption{The stability of each IQA. The difference is shown in relative percentage. The IQA is stable if the differences for scales and scores are small.}
  \label{fig:plot}
  \vspace{-5pt}
\end{figure}

We evaluate the stability of the deep IQAs.
We evaluate the stability in the image scale by computing the absolute differences between the default scale and the best scale.
The absolute differences are normalized by the best scale and represented by percentage.
We also evaluate the stability in the performance score by computing the absolute differences between the scores at the default scale and the best scale.
We show the results in Fig.~\ref{fig:plot}.
PieAPP is the most stable among the deep IQAs.

\begin{figure}[t]
  \centering
  \begin{minipage}[t]{0.48\hsize}
    \centering
    \includegraphics[width=\hsize]{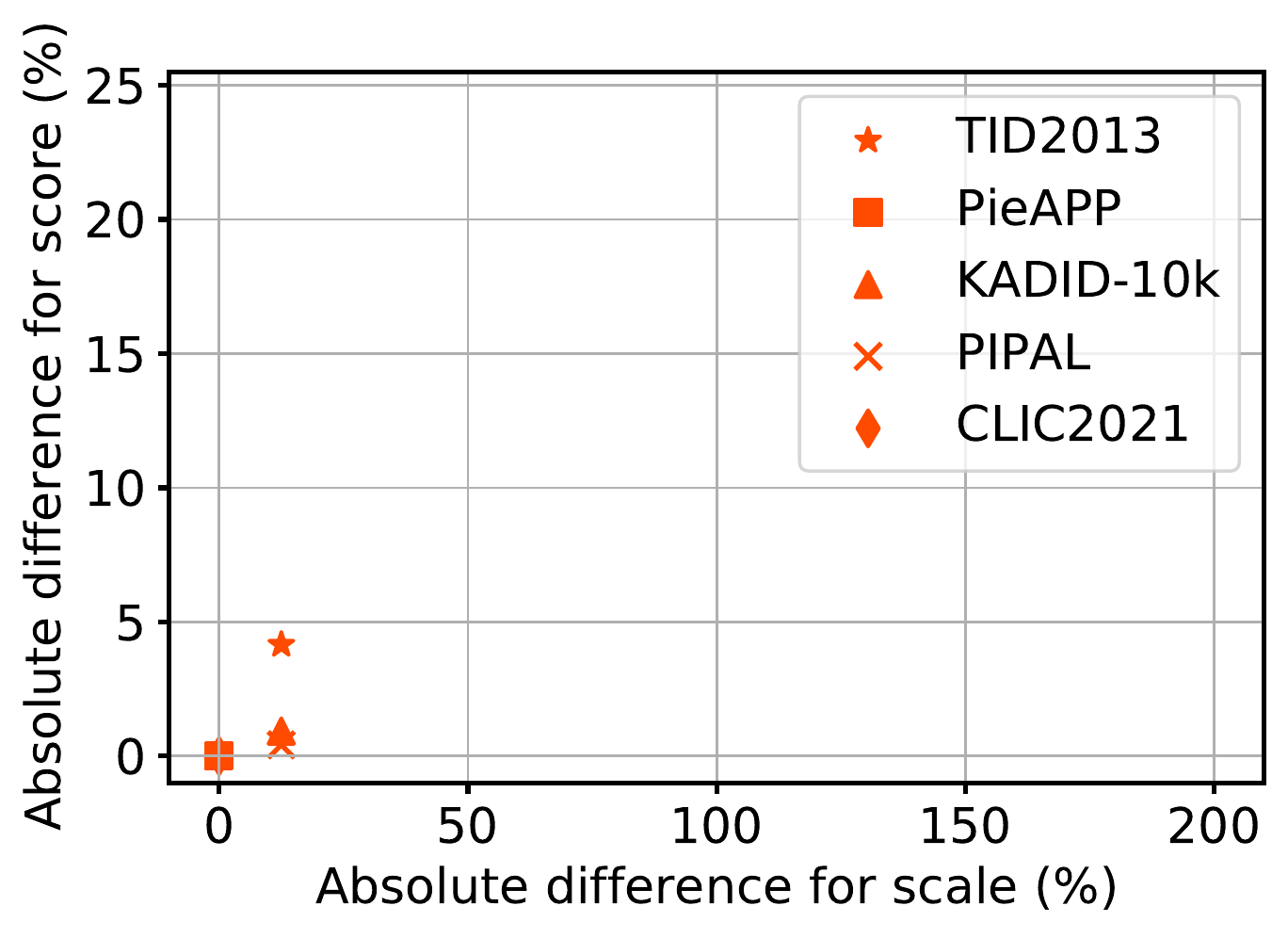}
    \subcaption*{LPIPS-RND-PieAPP}
  \end{minipage}
  \begin{minipage}[t]{0.48\hsize}
    \centering
    \includegraphics[width=\hsize]{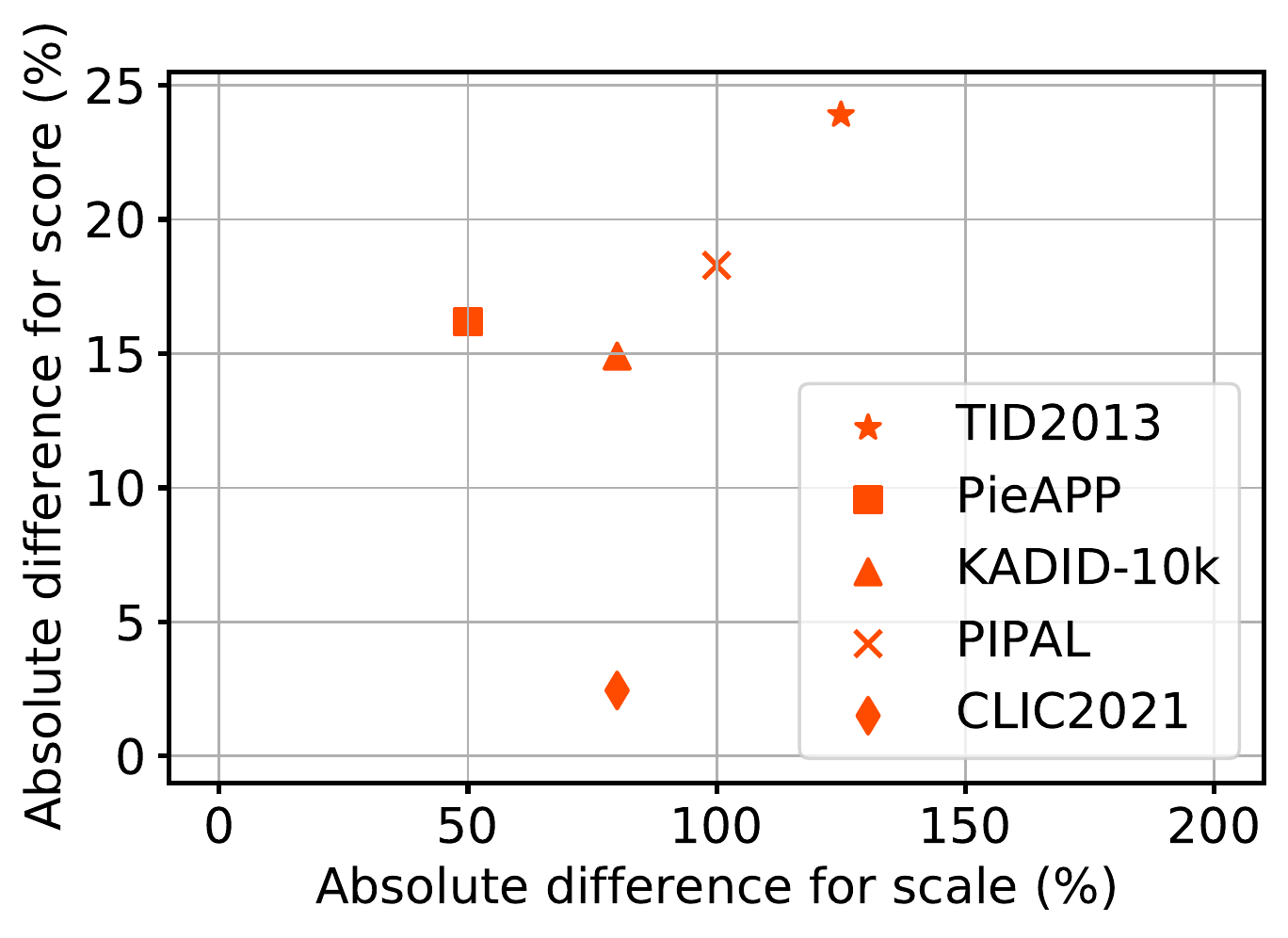}
    \subcaption*{LPIPS-RND-BAPPS}
  \end{minipage}
  \caption{The stability of LPIPS~\cite{LPIPS} with a backbone of AlexNet under different conditions. The difference is shown in relative percentage. The IQA is stable if the differences for scales and scores are small.}
  \label{fig:plot_abl}
  \vspace{-5pt}
\end{figure}

We consider the stability depends on the initial weights and the training data.
It is because these conditions are the main differences except the network architecture between PieAPP and the other three deep IQAs as shown in Table~\ref{tbl:method}.
To validate this, we add experiments of LPIPS under different conditions.
We train LPIPS with a backbone of AlexNet initialized with random weights on the PieAPP dataset (LPIPS-RND-PieAPP) and the BAPPS dataset (LPIPS-RND-BAPPS), respectively.
We show the results in Fig.~\ref{fig:plot_abl}.
LPIPS-RND-PieAPP is more stable than LPIPS-RND-BAPPS and LPIPS (Alex-lin) and competitive with PieAPP.
These results suggest that training deep IQAs on the PieAPP dataset from random initial weights contributes to the stability.

We show the qualitative results of the four IQAs on the TID2013 dataset in Fig.~\ref{fig:qualitative}.
The results demonstrate that the order of IQAs with the best scale is correct, whereas the order of IQAs with the default scale is incorrect.
This may be because the IQAs are sensitive to fine and sharp distortions, such as multiplicative Gaussian noise.

\section{Conclusions}
\vspace{-5pt}
In this paper, we have analyzed the effect of image scale on performance.
Experimental results demonstrated that down-scaling images has a large impact on performance and that the best image scale often differs from the default scale and the original size.
The best image scale differs depending on the model and the evaluation dataset.
We also found that PieAPP was the most stable among the deep IQAs.
Further investigation of the factor of stability is the future work.

\section*{Acknowledgements}
This work was partially supported by JSPS KAKENHI Grant Number 22J13735.

\bibliographystyle{ieicetr}
\bibliography{refs}


\end{document}